\RequirePackage{fix-cm}
\documentclass[twocolumn]{svjour3}

\smartqed
\usepackage{graphicx}
\usepackage{amsmath,amssymb}
\usepackage{hyperref}
\usepackage[linesnumbered, ruled, noend]{algorithm2e}
\usepackage{csquotes}
\usepackage{subcaption}
\usepackage{booktabs}

\usepackage{tikz}
\usetikzlibrary{positioning}
\usetikzlibrary{shapes,arrows}
\usepackage{pgfplots}
\pgfplotsset{compat=1.16}
\usepgfplotslibrary{fillbetween}

\usepackage{listings}
\usepackage{colortbl}
\colorlet{emph green}{green!50!black}
\colorlet{emph blue}{blue!90!black}
\colorlet{emph red}{red!75!black}
\colorlet{emph orange}{orange!75!black}
\colorlet{emph black}{black}

\graphicspath{ {images/} }

\journalname{German Journal of Artificial Intelligence}
\begin{document}
\title{Project Report: Requirements for a social robot as an information provider in the public sector}
\author{
	Thomas Sievers
	\and
	Nele Russwinkel
}
\authorrunning{Thomas Sievers and Nele Russwinkel}
\institute{
	Institute of Information Systems, University of Lübeck, \\ 23562 Lübeck, Germany \\
	\email{sievers@uni-luebeck.de, russwinkel@uni-luebeck.de} \\
}

\date{Received: date / Accepted: date}

\maketitle

\begin{abstract}
Is it possible to integrate a humanoid social robot into the work processes or customer care in an official environment, e.g. in municipal offices? If so, what could such an application scenario look like and what skills would the robot need to have when interacting with human customers? What are requirements for this kind of interactions? We have devised an application scenario for such a case, determined the necessary or desirable capabilities of the robot, developed a corresponding robot application and carried out initial tests and evaluations in a project together with the Kiel City Council. One of the most important insights gained in the project was that a humanoid robot with natural language processing capabilities based on large language models as well as human-like gestures and posture changes (animations) proved to be much more preferred by users compared to standard browser-based solutions on tablets for an information system in the City Council. Furthermore, we propose a connection of the ACT-R cognitive architecture with the robot, where an ACT-R model is used in interaction with the robot application to cognitively process and enhance a dialogue between human and robot.
\end{abstract}

\section{Application scenario and skills that a social robot should have in the public sector}

The idea of integrating a humanoid social robot into the work processes of public offices and authorities in a meaningful and beneficial way raises the question of a specific application scenario. In our scenario, the focus should be on contact with customers and the transfer of information. Specifically, we are talking about a deployment in the reception area of the residents' registration office of the city of Kiel. There is a need for information here, for example, on the type of documents required for a particular case or questions about any necessary registration and procedures. Employees of the office are involved in the design and testing of the robot applications with their own ideas and feedback.

In order to fulfill these and other tasks, the robot should be able to understand and respond in natural language. In a public authority environment, it would also be highly desirable if the robot could understand and speak foreign languages and carry out translations.

The information provided by the robot depends on the specific application and would have to be provided as "expert knowledge", for example from a database or stored in the application. Apart from answering questions in the respective specialist context, it would be advantageous for human-robot interaction to be as barrier-free and smooth as possible if the robot could understand and answer very general questions in addition to the specialist questions and, if necessary, conduct a dialogue freely.

In addition, it would be desirable if the information were not only provided in a national language that would be optimal for the City Council dialog partner, but also in easy language appropriate for the customer to facilitate access to important information. Easy language is defined as a language style with simpler sentence structure and additional explanations. The target groups that can benefit from information in easy language include, among others, people with little knowledge of the administrative language and people who are learning the official language as a second language. In Germany alone, more than 10 million people require text in easily understandable language to access crucial information and manage their daily lives \cite{easylanguage}.

The robot should also have basic social skills to provide sufficient incentives for the people it interacts with to accept its services. Apart from a suitable outward appearance, this includes an adequate response to human comments and also a way of dealing with errors or misunderstandings that prevents frustration and possible unwillingness on the part of humans as far as possible. In this context, it would be desirable if the robot would be able to recognize the current emotional state of the conversation partner and process it appropriately.

\section{Solution approaches for the desired tasks}

The technical approaches, architectures and services we use are listed below and provide an overview of the overall structure of the robot system to achieve the goals set in the project.

\subsection{Humanoid social robot Pepper}

\begin{figure*}
  \includegraphics[width=\textwidth]{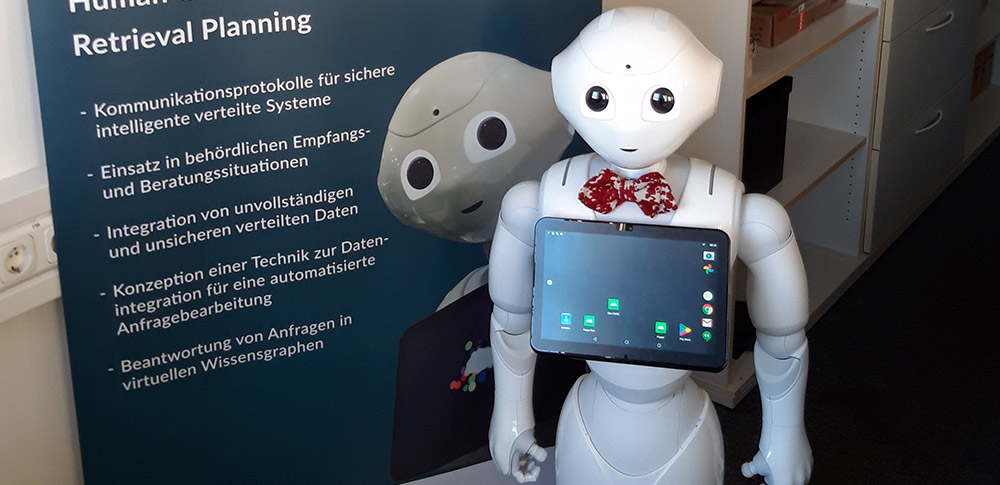}
  \caption{Social Humanoid Robot Pepper}
  \label{fig_pepper}
\end{figure*}

Since research has generally shown that trust is the basis for successful communication tasks and trust in robots is increased by anthropomorphism \cite{vanPinxteren}, we choose the humanoid social robot Pepper for the tasks to be accomplished. Pepper, shown in Figure~\ref{fig_pepper}, was developed by Aldebaran and first released in 2015 \cite{Pepper}. A human face, the possibility of human-like expressions and body language, and a name of its own are seen as beneficial for the trust of people in the robot. Pepper is 120 centimeters tall and optimized for human interaction. It is able to engage with people through conversation, gestures and its touch screen. It is equipped with internal sensors, four directional microphones in his head and speakers for voice output. The robot is able to process images from its 3D camera and two HD cameras by shape recognition software for identification of faces and objects so that it can focus on, identify, and recognize people. Speech recognition and dialogue are available in different languages, but the respective language package must be purchased for each language required. Beyond, Pepper can perceive basic human emotions. The robot features an open and fully programmable platform so that developers can program their own applications to be run on Pepper using software development kits (SDKs).

\subsection{Case-specific expert knowledge}

As a knowledge base for the case-specific expert knowledge that the robot should be able to provide information about, we use data from a MySQL database hosted on a web server with a self-programmed application programming interfaces (API) that returns text answers in JSON (JavaScript Object Notation) format \cite{10111181}. This API transmits a term or phrase that the robot has heard and recognized to the server via Wi-Fi connection. A PHP script running on the web server then returns a corresponding response from the database. This basic knowledge base could be something totally different in another case of application. It is also possible to store expected questions in the form of keywords or phrases together with answers directly in the application, so that an external data source is not necessarily required.

\subsection{Utilization of OpenAI GPT models}

In order to implement human-like arbitrary dialogues on any conceivable topic, we have had good experience with a connection to the OpenAI language models such as Generative Pre-trained Transformer (GPT) 3 or 4 \cite{openai}. Of course, the output of these language models is not deterministic, so their use harbors certain risks. However, by appropriately tuning the model parameters, using explicit prompt generation and fine-tuning the model, good results can be achieved, at least for a range of questions that are not subject-specific and where errors would not have serious consequences. In practice, there would certainly be some obstacles to the use of these language models in the public authority environment for reasons of data protection. However, a quasi-natural flow of dialogue between humans and robots can be achieved quite easily in this way.

\subsection{Translation into foreign languages and easy language}

For translation into a foreign language and into easy language, we use services that are hosted on external servers and are not proprietary. These services are also connected to the robot application via Wi-Fi and a corresponding API connection. We connect to the translation tool for easy language of SUMM AI and use the API of DeepL for translation services \cite{summ} \cite{deepl}. Text given to this APIs is automatically returned in a JSON format that is handed over to the voice and tablet output of the robot. SUMM AI currently supports easy language for German, English and some other languages. DeepL provides translation to 27 languages. The Pepper robot is equipped with the appropriate language package for being able to pronounce used languages correctly.

\subsection{Robot animation}

The use of robot gesture animation depending on a specific context is also possible with Pepper. Animations can be integrated depending on different events of the dialogue to enhance anthropomorphism and intelligibility through the indirect effect of body language. It is possible to define groups of appropriate animations, from which a randomly selected one is executed at certain points of the interaction, e.g., at the greeting, in response to a question from the human, when the robot asks a question, and so on. These animations support the interaction with the human as they underline statements of the robot. 

\subsection{Cognitive architecture ACT-R}

Interacting with humans in a natural way is a goal that humanoid social robots cannot easily achieve. In order to provide common ground for working towards a specific goal, and to flexibly react to actions of the partner and to develop situation understanding for adequate reactions, architectures such as ACT-R \cite{Anderson} can be used. Like any cognitive architecture, ACT-R as a theory for simulating and understanding human cognition aims to define the basic and irreducible cognitive and perceptual operations that enable the human mind. In theory, each task that humans can perform should consist of a series of these discrete operations. Most of the ACT-R's basic assumptions are also inspired by the progress of cognitive neuroscience, and ACT-R can be seen and described as a way of specifying how the brain itself is organized to produce cognition \cite{enwiki:1180065049}.

For our application area this cognitive architecture can provide flexible task knowledge and can build up mental representations of the relevant information that was provided regarding the individual the robot is working with or the state of the task that should be accomplished together. If, at some point, it turns out that the intention of the human partner can not directly be accomplished because some relevant information is missing, this person will probably be frustrated. Things like this need to be considered and appropriate reactions need to be retrieved by the robot to relate to this, e.g. find an alternative solution.

Pepper has the ability to recognize basic emotions via facial recognition, can pass these findings on to an ACT-R model, which in turn draws conclusions within the framework of the human-like cognitive architecture and controls the robot application via feedback. The ACT-R model thus controls, for example, the verbal reaction of the robot in the interaction with the human and adapts it to the emotion that has just been recognized. A combination of the possibilities of ACT-R with a humanoid social robot interacting directly with humans might be a way of making a conversation between a robot and a human more human-like on the part of the robot and therefore more pleasant for humans. 

\section{Experience so far}

Our project together with the Kiel City Council has been running since the end of 2022 over several stages, in which various adjustments have been made based on our experience in on-site testing and the ideas and wishes of the Kiel City Council. One general problem, for example, is the difficulty of using the robot to access data from the residents' registration office, as an initial idea in our scenario would have envisaged. This has not yet been feasible in the test phase of the project due to the municipality's IT security requirements, but could be simulated. However, there is agreement that the project will be continued and developed further. There is currently no defined end date.

We have developed various applications for the Pepper robot with which we have already been able to test different aspects of our ideas. In principle, the approaches outlined above work quite well. The translation of a text provided via the application from whatever source into another language functions properly. The same applies to conversion into easy language. Of course, translation and easy language can also be used in succession. With the exception of more complex requests to the OpenAI language models, there are no significant latencies. It is also possible to use DeepL to identify a language heard by Pepper, switch to the corresponding language pack and automatically respond in this language - provided the relevant language pack is available on the robot.

The robot itself is perceived as very likeable by almost everyone who meets it for the first time. We have had this experience both at Kiel City Council and at various events of different kinds. Children in particular are very interested and usually find it easy to get to know Pepper. However, there are great uncertainties in the assessment of the robot's abilities and often communication problems, which may be partly due to Pepper's limited vocabulary (when used without a large language model in the background), but also to the human dialogue partner's lack of knowledge of the modalities for addressing the robot appropriately. As a result, there is a risk of leaving an unfavorable impression on the human. This hazard must be minimized by clearly presenting the rules for effective communication, unless ways can generally be found to work through imponderables and avoid possible communication difficulties.

As far as the use of a cognitive architecture for controlling the application and modeling a more human-like interaction is concerned, it is a new research field to build situated human-aware agents interacting with human partners. For this a stepwise approach is necessary and we are at the very beginning of different cognitive concepts (e.g. situation understanding, predicting and adapting to the emotional state of the partner, flexible task anticipation) and the perception of human interaction partners.

The connection of an ACT-R application running on a PC or laptop to Pepper via the WLAN network was realized and an ACT-R model was created, which receives and processes the values transmitted by the robot's emotion recognition. Feedback from the model to Pepper controls its further behavior and the dialog. To evaluate an initial implementation of this concept, we have designed a study in which we want to compare a Pepper robot with a connection to the cognitive model with the robot without a cognitive model in order to test whether the human interaction partners had a more transparent impression of the interaction. The use of a more sophisticated ACT-R model in a scenario corresponding to our basic idea is still pending.

\section{Challenges}

The biggest challenge -- apart from the general difficulty of integrating quasi-external systems into an official IT infrastructure -- lies in the non-obvious nature of human-robot interaction, where most people do not realize what they can expect from a robot and how a conversation with the robot can be conducted in such a way that the robot understands what the human is saying. As soon as a dialog in natural language is possible, a human-like ability to understand is usually assumed, which is difficult to achieve in practice.

Due to a lack of experience in dealing with social robots, their capabilities are often overestimated. As already mentioned, this creates a great potential for frustration for the human interaction partner, which can significantly reduce the acceptance of the services offered by the robot.

Ambient noise or speech that is not directed at the robot but is picked up by it can contribute to misunderstandings or a complete lack of understanding. Pauses in human utterances and the usual timing of turn-taking also sometimes cause the robot to try to understand what has been said too early, even though the sentence has not yet been completed.

As timing plays an important role in a dialog, human conversation partners quickly become impatient if the robot takes too long to respond. This can happen especially when generating an answer with the help of a large language model such as GPT 3.5 or 4 from OpenAI if the answer is somewhat more complex and longer.

The Pepper robot is able to perceive people in its environment and approach them independently. However, an adequate social distance must be maintained, for which further tests and adjustments are required.

\section{Prospects}

Despite all the difficulties, the trials so far have been promising. However, the linchpin for successful use is likely to be an optimized and low-barrier access to the robot's capabilities and a scenario that favors easy access for humans. The robot must be enabled to interact smoothly with people who have no prior knowledge in this area. Specifically, the aim is to improve comprehension based on the Pepper robot's speech recognition or by other means, taking into account human expression habits such as pauses and filler words. And the robot should distinguish and express whether it has not understood what has been said acoustically or in terms of content.

It must be recognizable for the human conversation partner or become recognizable in the process which abilities the robot actually has - and preferably from the outset and not on the basis of previously failed communication attempts.  Fluid communication with an understanding of almost any topic on the part of the robot with the ability to always bring the conversation back to the topics specified by its task would be desirable. 

The use of large language models with their ability to generate human-sounding answers to almost any question is a good option here. Appropriate prompting as a starting point or instruction to the language model can be used, for example, to specify tonality and a specific framework. There is also the option of additional training and fine-tuning to provide the language model with special additional skills and knowledge.

It would also be conceivable to generate prompts for such a language model with the help of a cognitive architecture from an ACT-R model. This would combine human-like cognition with human-like language skills and could -- in combination with emotion recognition -- perhaps evoke something like empathic reactions on the part of the robot and make an interaction more pleasant for humans. We are looking forward to further developments and future results.

\bibliographystyle{splncs04}
\bibliography{literature.bib}

\end{document}